# DISCRIMINATION OF ENGLISH TO OTHER INDIAN LANGUAGES (KANNADA AND HINDI) FOR OCR SYSTEM

Ankit Kumar[1,] Tushar Patnaik[2] and Vivek Kr Verma[3]

[1,3]M.Tech Student, Department of Computer Science and Engineering, C-DAC Noida, INDIA

[1]ankitchaudhary6978@gmail.com
[3]vermavivek123@gmail.com

[2] Sr. Lecturer/Sr. Project Engineer, Department of Computer Science and Engineering, C-DAC Noida, INDIA

[2]tusharpatnaik@cdac.in

## ABSTRACT

*India is a multilingual multi-script country. In every state of India there are two languages one is state local language and the other is English. For example in Andhra Pradesh, a state in India, the document may contain text words in English and Telugu script. For Optical Character Recognition (OCR) of such a bilingual document, it is necessary to identify the script before feeding the text words to the OCRs of individual scripts. In this paper, we are introducing a simple and efficient technique of script identification for Kannada, English and Hindi text words of a printed document. The proposed approach is based on the horizontal and vertical projection profile for the discrimination of the three scripts. The feature extraction is done based on the horizontal projection profile of each text words. We analysed 700 different words of Kannada, English and Hindi in order to extract the discrimination features and for the development of knowledge base. We use the horizontal projection profile of each text word and based on the horizontal projection profile we extract the appropriate features. The proposed system is tested on 100 different document images containing more than 1000 text words of each script and a classification rate of 98.25%, 99.25% and 98.87% is achieved for Kannada, English and Hindi respectively.*

## Keywords

## 1. Introduction

In Multilingual document analysis, it is important to automatically identify the scripts before feeding each text words of the document to the respective OCR system. In India, English has proven to be the binding language. Therefore, a bilingual document page may contain text words in regional language and English. So, bilingual OCR is needed to read these documents. To make a bilingual OCR successful, it is necessary to separate portions of different script regions of the bilingual document at word level and then identify the different script forms before running an individual OCR system.

In the context of Indian language document analysis, major literature is due to Pal and Choudhari. The automatic separation of text lines from multi-script documents by extracting the features from profiles, water reservoir concepts [1]. Santanu Choudhury, Gaurav Harit, Shekar Madnani and R. B. Shet has proposed a method for identification of Indian languages by combining Gabor filter based technique and direction distance histogram classifier considering Hindi, English, Malayalam, Bengali, Telugu and Urdu [2]. Chanda and Pal have proposed an automatic technique for word wise identification of Devnagari, English and Urdu scripts from a single document [3]. Word level script identification in bilingual documents through discriminating features has been developed by B V Dhandra, Mallikarjun Hangarge, Ravindra Hegadi and V.S.Malemath [4].Vijaya and Padma has developed methods for English, Hindi and Kannada script identification using discriminating features and top and bottom profile based features (English, Hindi, Kannada) [5]. B.V.Dhandra, H.Mallikarjun, Ravindra Hegadi, V.S.Malemath developed a method of Word-wise Script Identification from Bilingual Documents Based on Morphological Reconstruction(English, Hindi, kannada) [6]. Prakash K. Aithal, Rajesh G., Dinesh U. Acharya, Krishnamoorthi M. Subbareddy N. V. Has proposed a method of Text Line Script Identification for a Multilingual Document (English, Hindi, Kannada) [7].

This paper deals with word-wise script identification for Kannada, English and Hindi script pertaining documents from Karnataka, Uttar Pradesh. Script identification is done based on the features extracted from Horizontal Projection Profile and the vertical projection profile of the word segment. To discriminate Kannada, English and Hindi the mean of horizontal Projection Profile Values between first and second largest and value of the point immediately after either first largest or second largest depending upon the position, which largest come earlier in the horizontal projection profile is used.

Secondly after calculating the above feature, we calculate the vertical strokes present in the word in order to achieve better accuracy in the result. After analysing the all three script we see that Hindi and English language contain vertical strokes in their words (for example-B, D, E, T, I, d, b, n, k, etc in English). But in case of Kannada vertical strokes is not present. For vertical strokes first we calculate the height of every word by using horizontal projection, after that by using vertical projection we calculate the vertical strokes equal to the word height. These strokes are considered for feature extraction.

## 2. RESEARCH METHODOLOGY

### 2.1 Discriminating Features of Hindi, English and Kannada

**1.** In English script vertical strokes appear in the left side of the character mostly such as (B, D, H, F, R, K, P, b, h, k, l) whereas in Hindi they appear in the right side of the characters as shown in the fig 1.

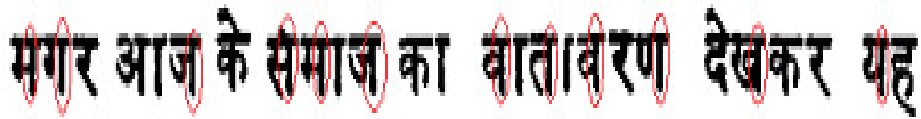

Figure 1 Vertical stroke in the right side of the characters

**2**. In most cases height of English character is greater than its width, whereas in Kannada width of character is greater than height as shown in fig 2(a).
**3**. Horizontal stroke is more in Kannada script compare to the English script shown in fig 2(b).
**4**. Aspect ratio of English character is greater than 1, whereas in case of Kannada it is less than 1.it is the ratio of height and width.

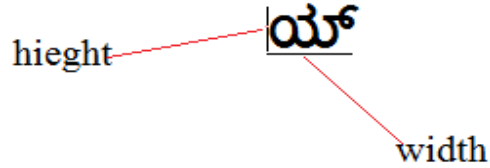

Figure 2(a) Height and width of a Kannada character

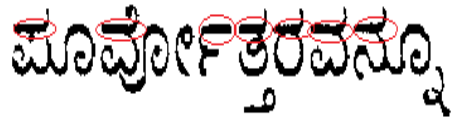

Figure 2(b) Horizontal stroke in Kannada words

**5.** Most of the Hindi language characters (alphabets) have a horizontal line at the upper part. In Hindi, this line is called sirorekha as shown in fig 3.

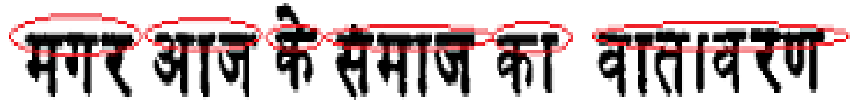

Figure 3 Headline at the upper part in the Hindi script

## 2.2 Pre-processing

The documents are scanned using HP Scanner. The scanned images are digitized images and are in gray tone. We have to convert the image into the binary image. For this we use image binarization. After binarization we perform skew detection and skew correction on the binary image.

### 2.2.1 Binarization

 System takes input image in gray tone having pixels intensity values between   (0-255) and using a thresh-holding approach converts them into two-tone images (0 and 1), black pixels having the value 1's correspond to object and white pixels having value 0's correspond to background.

### 2.2.2 Dilation and Erosion

When we perform word level segmentation on the binarize document then we do not get good accuracy due to broken character or word. In that case a single word is segmented into more than one word. We perform dilation and erosion on the document in order to remove this deficiency. In fig 3(a) original image is shown and in fig 3(b) image after dilation and erosion is given.

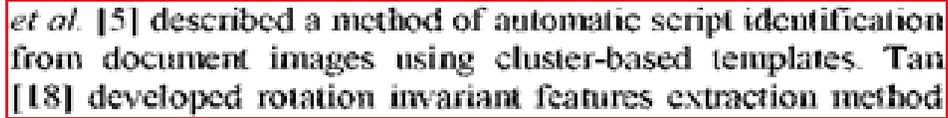

Figure 3(a) Binarize image document.

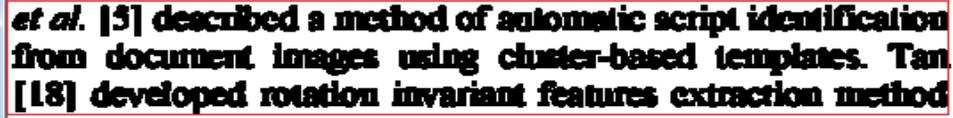

Figure 3(b) Image document after dilation and erosion.

On the resultant images when we perform word level segmentation, we get the exact positions of word start and end.

### 2.2.3 Segmentation

White space between text lines is used to segment the text lines. The line segmentation is carried out by calculating the horizontal projection profile of the whole document. The horizontal projection profile is the histogram of number of black pixels along every row of the image. The projection profile exhibits valleys of zero height corresponding to white space between the text lines. Line segmentation is done at these points.

Similarly White space between text words is used to segment the text lines. Word segmentation is done by the vertical projection profile. The vertical projection profile is the histogram of number of black pixels along every column of the segmented line. The projection profile exhibits valleys of zero height corresponding to white space between the text words.

## 3. PROPOSED WORK

This paper deals with word-wise script identification for Kannada, English and Hindi script pertaining documents from Karnataka, Uttar Pradesh. Script identification is done based on the features extracted from Horizontal Projection Profile and the vertical projection profile of the word segment. To discriminate Kannada, English and Hindi the mean of horizontal Projection Profile Values between first and second largest and value of the point immediately after either first largest or second largest depending upon the position, which largest come earlier in the horizontal projection profile is used.

Secondly after calculating the above feature, we calculate the vertical strokes present in the word in order to achieve better accuracy in the result. After analysing the all three script we see that Hindi and English language contain vertical strokes in their words (for example-B, D, E, T, I, d, b, n, k, etc in English). But in case of Kannada vertical strokes is not present. For vertical strokes first we calculate the height of every word by using horizontal projection, after that by using vertical projection we calculate the vertical strokes equal to the word height. These strokes are considered for feature extraction.

### 3.1 Algorithm

1) For each word calculate the word height (Wh) by using horizontal projection profile.
2) Find the number of vertical strokes (Vs) equal to the word height (Wh) by using vertical projection profile.

3) For each word calculate the first (L1) and second (L2) largest value of the horizontal projection profile.
4) Calculate the Largest mean, Lm (Largest mean is the mean of projection profile between the first and second Largest including both).
5) Find the value of the point Lp (Lp is the point immediately after the Largest (L1orL2) which come first in the horizontal projection profile).
6) Compare the Lp with Lm and find number of Vs:-
   (a) If Lp/Lm falls in the range 0.071-0.258 and Vs >1 then the text word is recognized as Hindi.
   (b) Else If Lp/Lm falls in the range 0.258- 0.5 and Vs <=1 then the text word is recognized as Kannada.
   (c) Else if Lp/Lm falls in the range 0.5-0.9 and Vs >1 then the text word is recognized as English.

## 4. RESULTS

The database includes 700 text words from 30 different document images. The document images are downloaded from Google and e-news paper (Times of India, Hindustan times, Sanjevani, vijaya Karnataka, Amar Ujala) for English, Hindi and Kannada. It includes the text words both in regular and italics fonts of size varying from 9 to 14. The proposed system is tested on 100 different document images containing 1000 text words of each script and a classification rate of 96.25%, 99.25% and 98.87% is achieved for Kannada and English and Hindi respectively.

Table 1. Range of Lp/Lm for Kannada, English and Hindi

| Language | Range |
|----------|-------|
| Kannada  | 0.31 to 0.50 |
| English  | 0.5 to 0.96 |
| Hindi    | 0.071-0.31 |

In our approach in case of italic fonts vertical strokes is not calculated correctly but in testing we see that all italics words is classified correctly on the basis of the range of Lp/Lm.

Table 2. Identification results of Kannada and English

|         | Kannada | English |
|---------|---------|---------|
| Kannada | 98.25%  | 1.75%   |
| English | 1.25%   | 98.75%  |

Table 3. Identification results of Hindi and English

|  | Hindi | English |
|---|---|---|
| Hindi | 99.65% | 0.35% |
| English | 0.33% | 99.67% |

Table 4. Confusion matrix of script Identification (Tri-lingual Document)

|  | Kannada | English | Hindi |
|---|---|---|---|
| Kannada | 98.25% | 0.5% | 1.25% |
| English | 0.725% | 99.25% | 0.025% |
| Hindi | 0.24% | 0.885% | 98.875% |

Table I show the range of Lp/Lm for Kannada, English and Hindi text words. Table II gives the identification results of Kannada and English. Table III gives the identification results of Hindi and English. Table IV gives the confusion matrix for the proposed script identification.

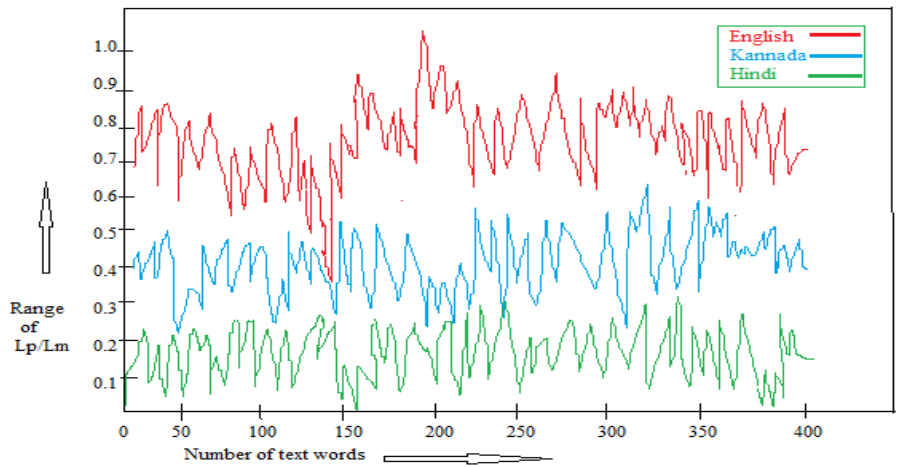

Figure 4. Range of Lp/Lm for Hindi, English and Kannada script

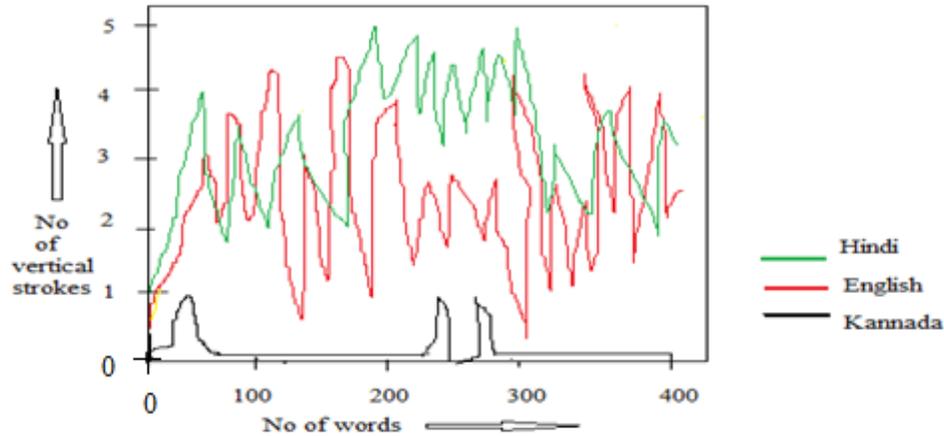

Figure 5.Appearance of number of vertical strokes in each script.

Fig.4 shows the range of Lp/Lm for Hindi, English and Kannada scripts. Fig.5 shows appearance of number of vertical strokes in each script. In Fig.5 we can see that vertical strokes in Kannada language are very less compare to the other two languages. In Hindi script vertical strokes is maximum.

## 5. CONCLUSION

In this paper, a simple and efficient algorithm for script identification of Kannada, English and Hindi text words from printed documents is proposed. The approach is based on the analysis of horizontal and vertical projection profile and does not require any character segmentation. It is based on word level segmentation. The system exhibits an overall accuracy of 98.792%.

## 6. COMPARISON OF PROPOSED WORK TO OTHER EXISTING METHODOLOGY.

**1.** In Morphological Reconstruction based approach an automatic technique for script identification at word level is proposed for two printed bilingual documents of Kannada and Devnagari containing English numerals (printed and handwritten)[5]. But in our work both character and numerals of English script is identified correctly.
**2.** In Discriminating feature based approach identification is done based on eccentricity, aspect ratio, strokes of the word segmented and approach work for English numerals not for characters [3].
In our work identification is done based on feature extracted from horizontal projection profile which enhances the accuracy and simplicity of the algorithm.
**3.** Many script Identification algorithm exists for these three languages in which text line is identified[4][6], but we know that within a line text words of many languages can appear due to this accuracy of algorithm decreases. In our work identification is done on word level.

Table 5. Comparison of proposed work to other existing methodology

| Existing Methodology | Dataset | Overall accuracy |
| --- | --- | --- |
| Morphological Reconstruction[5] | 2250(words) | 96.10% |
| Discriminating Feature[3] | 2500(words) | 95.85% |
| Top-profile and Bottom-profile[4] | 600(text lines) | 96.6% |
| Feature extracted from text line Horizontal Projection[6] | 450(text lines) | 99.83% |
| Proposed Algorithm | 100 documents images(more than 3500 words) | 98.792% |

## 7. FUTURE WORK

The work could be extended to character level script identification and for other Indian scripts. We can also introduce scaling and thinning algorithms in current algorithm so that the font size dependency can be removed. After applying thinning and smoothing our approach will work on boldface also.

## ACKNOWLEDGEMENT

The authors are thankful to the referees for their critical comments. We also thankful to Bhupendra Singh (Sr. Scientific officer) and Mr. Deepak Kumar Arya (Engineer IT) at C - DAC Noida, for his helpful suggestions.

**Authors:**

Mr. Ankit Kumar received his B.Tech in Computer Science and Engineering from U.P.T.U Lucknow, Uttar Pradesh, India in 2010.Currently, he is doing M.Tech in Computer Science and Engineering from C-DAC Noida(Affiliated to G.G.S.I.P.U New Delhi), India. His interest areas are Digital Image Processing, Network Security, Theory of Computation, Data Structure and Data Base Management System. 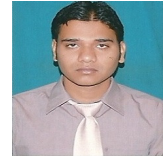

Mr. Tushar Patnaik (Sr. Lecturer/Sr. Project Engineer) joined CDAC in 1998. He has eleven years of teaching experience. His interest areas are Computer Graphics, Multimedia, Database Management System and Pattern Recognition. At present he is leading the consortium based project "Development of Robust Document Analysis and Recognition System for Printed Indian Scripts". 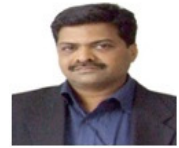

Mr. Vivek Kr.Verma received his B.Tech in Computer Science and Engineering from R.T.U Kota, Rajsthan, India in 2010.Currently, he is doing M.Tech in Computer Science and Engineering from C-DAC Noida(Affiliated to G.G.S.I.P.U New Delhi), India. His interest areas are Digital Image Processing, Computer Graphics, Compiler Design and Data Structures. 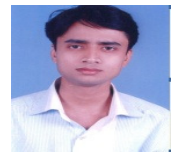